\documentclass{article}

\usepackage{microtype}
\usepackage{graphicx}
\usepackage{subfigure}
\usepackage{booktabs} % for professional tables

\usepackage{hyperref}
\usepackage{fourier}

\usepackage{amsmath}
\usepackage{amssymb}
\usepackage{mathtools}
\usepackage{amsthm}

\usepackage[capitalize,noabbrev]{cleveref}

\theoremstyle{plain}
\newtheorem{theorem}{Theorem}[section]

\theoremstyle{definition}
\newtheorem{definition}[theorem]{Definition}

\theoremstyle{remark}

\usepackage[disable,textsize=tiny]{todonotes}
\usepackage{enumitem}
\usepackage{multirow}
\usepackage{listings}
\usepackage{xcolor}

\usepackage{lscape}

\begin{document}

\title{Generative Modeling of Complex Data}

\author{
  Luca Canale\\
  \href{mailto:lc@sarus.tech}{lc@sarus.tech}
  \and
  Nicolas Grislain\\
  \href{mailto:ng@sarus.tech}{ng@sarus.tech}
  \and
  Johan Leduc\\
  \href{mailto:jl@sarus.tech}{jl@sarus.tech}
  \and
  Grégoire Lothe\\
  \href{mailto:gregoire.lothe@polytechnique.edu}{gl@sarus.tech}
}

\maketitle

\begin{abstract}
In recent years, several models have improved the capacity to generate synthetic tabular datasets.
However, such models focus on synthesizing simple columnar tables and are not useable on real-life data with complex structures.
This paper puts forward a generic framework to synthesize more complex data structures with composite and nested types.
It then proposes one practical implementation, built with causal transformers,
for \emph{struct} (mappings of types) and \emph{lists} (repeated instances of a type).
The results on standard benchmark datasets show that such implementation consistently outperforms
current state-of-the-art models both in terms of machine learning utility and statistical similarity.
Moreover, it shows very strong results on two complex hierarchical datasets with multiple nesting and sparse data, that were previously out of reach.
\end{abstract}

\section{Introduction}
\label{intro}

Data has become a critical resource for every sector of the economy. At the same time, new risks for privacy have emerged and regulators have been setting new legal standards such as HIPAA, GDPR, CCPA or PIPL to mitigate them. In particular, these regulations increasingly recognise synthetic data as a valid option for entities to exchange and analyse personal data while preserving both privacy and most of the statistical properties of the original data.

Surprisingly,  not so many studies focus on tabular synthetic generation: traditional techniques include decision trees,
bayesian networks or copulas.  In the last years, the success of \emph{Generative Adversarial Networks} \cite{goodfellow2014generative} in Computer Vision has also pushed towards
the development of such models for tabular data generation. However,
all these models (traditional and GANS), only address the problem of a \emph{single} table.

To our knowledge, the case of multiple relational tables (or hierarchical datasets) has little litterature.
We could find one systematic approach \cite{Patki2016SyntheticDataVault} based on gaussian copulas that we will later develop more. Such approach however does not scale well and yields relatively poor results.

Yet, in real-life use cases, data almost never come in a nice and clean tabular format,
but more as relational data across many tables or from so called \emph{data lakes} in complex
format such as \emph{apache parquet} \cite{parquet} or \emph{avro} \cite{avro}.
These data cannot be easily represented in a standard columnar fashion with common types (floats, strings..) but rather in hierarchical/composite structures, where types may be optional, repeated, or even be structured themselves into sub-types. Hence, synthesizing such data is still challenging today and often performed manually.

In this paper, we introduce a new formalism that maps, to a large class of composite types,
a generative model or \emph{codec}. Exactly as types, codecs can be composed to form composite codecs,
so that generating structured data can be achieved systematically.
%\begin{itemize}[topsep=2pt,itemsep=2pt,partopsep=2pt, parsep=2pt]

\textbf{Contributions}. We introduce a new framework that systematically maps a large class of data types to generative models called \emph{codecs}. We explicit the codecs architecture for primitive types (categorical and numerical data),
and composite (structs and lists) and show that in particular these codecs allow us to synthesize standard and complex hierachical tabular datasets.

We show that, on simple benchmarks, our model consistently outperforms state-of-the-art models. We also show that
for hierarchical datasets, the synthetic samples preserve statistical properties well and that
a machine learning model trained on synthetic data yields good performances on the original data.
%\end{itemize}

\section{Related work}

\paragraph{Tabular data generation}
Generation of synthetic tabular data remains a challenge for deep learning.
Classical approaches include copulas \cite{Patki2016SyntheticDataVault,li2020Sync} and bayesian networks \cite{zhang2017PrivBayes}.

Deep learning approaches like GANs \cite{goodfellow2014generative} have achieved great performances on homogeneous data such as audio and images.
They have faced difficulties on tabular data due to the nature of categorical features.
TableGAN \cite{park2018data} first applied the DCGAN architecture to tabular data but introduced spatial inductive bias in the process.
MedGAN \cite{camino2018generating} generates medical records but can only be applied to categorical data and does not generalize to other tabular types.
CT-GAN \cite{xu2019modeling} introduced a mode specific encoding of real features,
used the Gumbel softmax trick to backpropagate categorical sampling, and masks to counterbalance mode collapse. The state of the art, CTAB-GAN improves the way the conditioning is done and better handles mixed columns \cite{zhao21}.

Variational Autoencoders (VAE) like TVAE \cite{xu2019modeling} and VAEM \cite{ma2018vaem} as well as learned structured causal models
\cite{wen2022causaltgan} and invertible flows \cite{kamthe2021copula} have also been proposed for the task.

\paragraph{Transformers on tabular data}
The transformer architecture \cite{vaswani2017attention} has been applied to tabular data for machine learning tasks.
TabTransformer \cite{Huang2020TabTransformerTD} obtained good performances by embedding the categorical features and producing contextual embeddings through a transformer.
FT-Transformer \cite{gorishniy2021revisiting} tokenized both numerical and categorical features and used a class token to aggregate the contextual embeddings. TabularGPT \cite{Padhi2021TabularGPT} synthesizes multivariate time series,  each time series being modeled by one GPT-2 like transformer.
\paragraph{Hierarchical datasets} Altough many works have been published on tabular data,
surprisingly few tackle the more harduous task of generating hierarchical datasets.
Synthetic data vault \cite{Patki2016SyntheticDataVault}, or HMA1 \footnote{\url{https://sdv.dev/SDV/user_guides/relational/hma1.html}} is to our knowledge the only algorithm to generate relational databases.
It first extends parent tables with distribution statistics of referring rows and then uses one copula to model each table.

%\paragraph{Assessment of synthetic data quality} The most common approach is \emph{machine learning efficacy}
%\cite{choi2017generating,xu2019modeling}.
%It consists in comparing the performance of a classifier trained on the synthetic dataset with the same classifier trained on the real dataset.
%A visual approach consists in comparing directly the marginals \cite{camino2018generating,choi2017generating}
%while a more quantitative approach consists in using statistical tests, such as the Kolmogorov-Smirnov test.
%Some authors \cite{bowen2019comparative}, sometime compare 2-way, 3-way or higher order marginals.
%
%\todo[author=LC,inline]{Remove this last paragraph}

\section{Composite Generative Models}
% Can be ralated to https://en.wikipedia.org/wiki/Algebraic_data_type

% nommages:
% embedding vector e,
% intermediate context h,
% conditioning vector c,
% distribution representation d
% peut etre alleger les repetitions (E,D,C,L) pour tous les types

We are interested in modeling the distribution of $x \in  \mathfrak{X}$,
given a data set $X = \left(x_1, x_2,\ldots, x_N\right) \in \mathfrak{X}^N$.
The set $\mathfrak{X}$ of possible $x$ will be defined as the values of some,
potentially complex \emph{type}.

\subsection{Abstract Generative Model}

\subsubsection{Model Definition}

A generative model is a statistical model $C_\theta$,
that can be fitted to some observations $x \in \mathfrak{X}$
assumed to be i.i.d. draws from some distribution,
and can be used to generate new observations in $\mathfrak{X}$.

\begin{definition}
A \emph{codec} is a generative model defined by a quadruplet:
$C = \left( E, D, S, L\right)$
and a fixed \emph{initial conditioning vector} $c_0 \in \mathcal{E}$. Namely:

\paragraph{An encoder}
$$E : \mathfrak{X} \to \left(\mathcal{E} \times \mathcal{H}\right)$$

The encoder takes an observation $x\in\mathfrak{X}$
and returns a pair $(e,h) \in \left(\mathcal{E} \times \mathcal{H}\right)$ of $e \in \mathcal{E}$
an \emph{embedding vector} summarizing everything there is to know about $x$, and $h\in\mathcal{H}$ an
\emph{intermediate context} containing all intermediate information about a composite observation while being processed ---
this is explicited in \ref{sec:composite}.
For simple codecs, where the intermediate context is not used, instead of using a degenerate
$\mathcal{H}=\{\bullet\}$ with a single element and note $(e,\bullet) \in \left(\mathcal{E} \times \{\bullet\}\right)$,
we will simply note $e \in \mathcal{E}$.

\paragraph{A decoder}
$$D : \left(\mathcal{E} \times \mathcal{H}\right) \to \mathcal{D}\left(\mathfrak{X}\right)$$
The decoder takes a \emph{conditioning vector} $c \in \mathcal{E}$ from the embedding space
and an intermediate context $h\in\mathcal{H}$, and returns $d\in \mathcal{D}\left(\mathfrak{X}\right)$ a
\emph{distribution representation}.

\paragraph{A sampler}
$$S : \left(\mathcal{E} \times \Omega\right) \to \mathfrak{X}$$
Given some probability space $(\Omega ,\mathcal{F},P)$,
the sampler is a random variable taking a \emph{conditioning vector} $c \in \mathcal{E}$,
an outcome $\omega \in \Omega$ and returning a sampled observation $x \in \mathfrak{X}$.

\paragraph{A loss function}
$$L : \left(\mathcal{D}\left(\mathfrak{X}\right) \times \mathfrak{X} \times \Omega\right) \to \mathbb{R}$$
The loss function is a measurable function taking a decoded \emph{distribution representation}, an actual observation, an
outcome $\omega \in \Omega$ and returning a loss (e.g. negative log-likelihood).
Note that the loss function is a random variable, unless $\omega$ is ignored.

In the context of generative modeling, the loss function will usually be related to negative log-likelihood (or cross-entropy)
or an approximation of it.

\end{definition}

A codec is usually parametrized so that it can be fitted to real data:
$$C_\theta = \left( E_\theta, D_\theta, S_\theta, L_\theta\right)$$.
Please note that parameters may be shared between the various components of the codec.

\begin{figure}[!htb]
\begin{center}
\centerline{\includegraphics[width=\columnwidth]{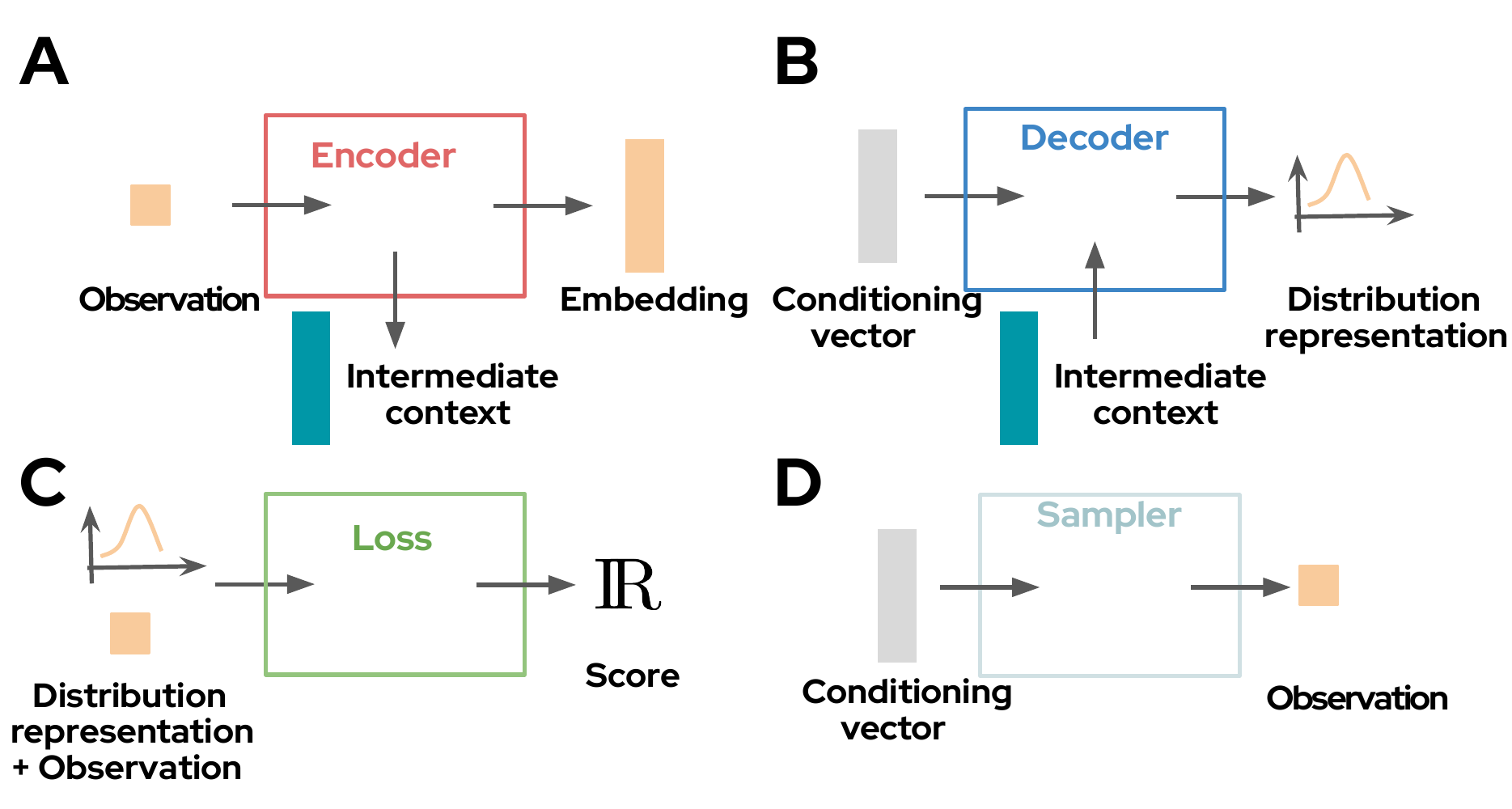}}
\caption{Sketch of the four attributes that characterize a generic codec: an encoder (\textbf{A}), yielding an embedding and an intermediate context from an observation;
a decoder (\textbf{B}) returning a distribution representation from a conditioning vector and an intermediate context;
a loss function (\textbf{C}) evaluating a score between an observation and a distribution representation; and a sampler (\textbf{D}) that returns an observation from a conditioning vector.}
\label{codec_abstract}
\end{center}
\end{figure}

\subsubsection{Training Process}

To fit $C_\theta$ to some data $X = \left(x_1, x_2,\ldots, x_N\right) \in \mathfrak{X}^N$,
the observations are encoded into $(e_i, h_i) = E_\theta(x_i)$, they are then decoded using the
\emph{initial conditioning vector}
into a \emph{distribution representation} $d_i = D_\theta(c_0, h_i)$
that is used to compute the loss $l_i(\omega) = L_\theta(d_i,x_i,\omega)$. The empirical loss on the
batch of data is given by:
$$\mathcal{L}(\theta, \omega) = \frac{1}{N}\sum_{i=1}^N L_\theta\left(D_\theta\left(c_0, E_\theta(x_i)_2\right),x_i,\omega\right)$$
with $E_\theta(x_i)_2 = h_i$ the second element in the pair $E_\theta(x_i) = (e_i, h_i)$.
$\mathcal{L}(\theta,\omega)$ can be minimized by regular Stochastic Gradient Descent (SGD).

This process may seem overly complex, since the embeddings $e_i=E_\theta (x_i)$ computed for each observation are not used,
but this formalism is in fact very powerful when generative models are combined together, as in \ref{sec:composite}.

\subsubsection{Sampling Process}

The sampling process is straightforward:
$$x \sim S_\theta(c_0)$$
the \emph{sampler} is fed with the \emph{initial conditioning vector}
and outputs a new synthetic observation.

\subsection{Primitive Data Types and Codecs}

We first define codecs for primitive types: enumerations, numbers, etc.
These primitive codecs are the building blocks that will be composed together into more complex types.

\subsubsection{Categorical Codec}

An observation $x$ of \emph{categorical type} is an element of a finite set $\mathfrak{X} = \left\{a_1,a_2,\ldots, a_n\right\}$ of cardinality $n$.

A \emph{categorical codec} $C_\text{cat}$ is one of the simplest generative model imaginable.

The encoder:
$$E_\text{cat} = a_k \mapsto W_k$$
simply encodes a value $x=a_k$ into an embedding vector $W_k \in \mathcal{E} = \mathbb{R}^d$.
It is parametrized by the embedding matrix $W \in R^{n\times d}$.

The decoder:
$$D_\text{cat} = c \mapsto c \cdot W^T$$
projects a conditioning vector onto the embedding vectors to obtain
a distribution representation in the form of an un-normalized vector of log-probabilities,
or \emph{logits}.

The sampler draws observations from the discrete distribution represented by the logits.
$$S_\text{cat} = c \mapsto x \sim \text{softmax}\left(c \cdot W^T\right)$$

The loss $L_\text{cat}$ for one observation $a_k$ is the categorical cross-entropy with the decoded distribution representation $d$.
$$L_\text{cat} = (d, a_k) \mapsto \ln\left[\text{softmax}\left(d\right)\right]_k$$
The loss of a batch of observations is the empirical mean of the observation losses.

\subsubsection{Numerical Codec}

An observation $x$ of \emph{numerical type} is an element of $\mathfrak{X} \subset \mathbb{R}$ (or $\mathbb{Z}$). \emph{Numerical codec}s treat numerical observations as categorical via a pre-processing
that consists in estimating a certain number $n$ of quantiles $q_i$ for the underlying distribution
and then assigning to each value its closest quantile.
Then the value can be treated as a categorical for the encoding.
Conversely, for decoding, the categorical sampled is an integer $i \in \left \lbrace 1,n \right \rbrace$ and we can sample uniformly between $q_i$ and $q_{i+1}$.

\subsection{Composite Data Types and Composite Codecs}
\label{sec:composite}

Composite data types are types built out of other types: primitive types or other composite types, such as:
\begin{itemize}
  \item \emph{Product types} or \emph{Structs} containing observations built out of many sub-observations --- or features --- of various types.
  \item \emph{Sum types} or \emph{Tagged Unions} containing observations of one of the many sub-types.
  \item \emph{Lists} containing finite sequences of observations of a given type.
\end{itemize}
Such composite types enable the construction of many complex types found in real-life data analysis.

We can show that similar constructs, called \emph{Composite Codecs}
can be defined to combine complex generative models out of simpler ones.
Those \emph{Codecs} use the \emph{chain rule} of probability:
$$\mathrm {P} (x_{n},\ldots,x_{1}) = \prod_{k=1}^n \mathrm {P} (x_{k}\mid x_{k-1},\ldots,x_{1})$$
If $x_{1},\ldots,x_{k}$ is summarized by some embedding vector $e_k$, we have:
$$\mathrm {P} (x_{n},\ldots,x_{1}) \simeq \prod_{k=1}^n \mathrm {P} (x_{k}\mid e_{k-1})$$
and it is easy to see how the generation of composite types reduces to the chaining of many
encoders and conditional samplers. The following sections detail two implementations of such composite codecs:
the ones for the \emph{struct} and \emph{list} types.

\subsubsection{Struct Codec}

An observation $x$ of \emph{struct type} is a tuple of features $\left(x_1, x_2,\ldots, x_n\right)$,
associated with labels. Here we assume the labels are simply the indices $1,2,\ldots,n$.
Each feature has its own type.

\begin{figure}[!ht]
\vskip 0.2in
\begin{center}
\centerline{\includegraphics[width=\columnwidth]{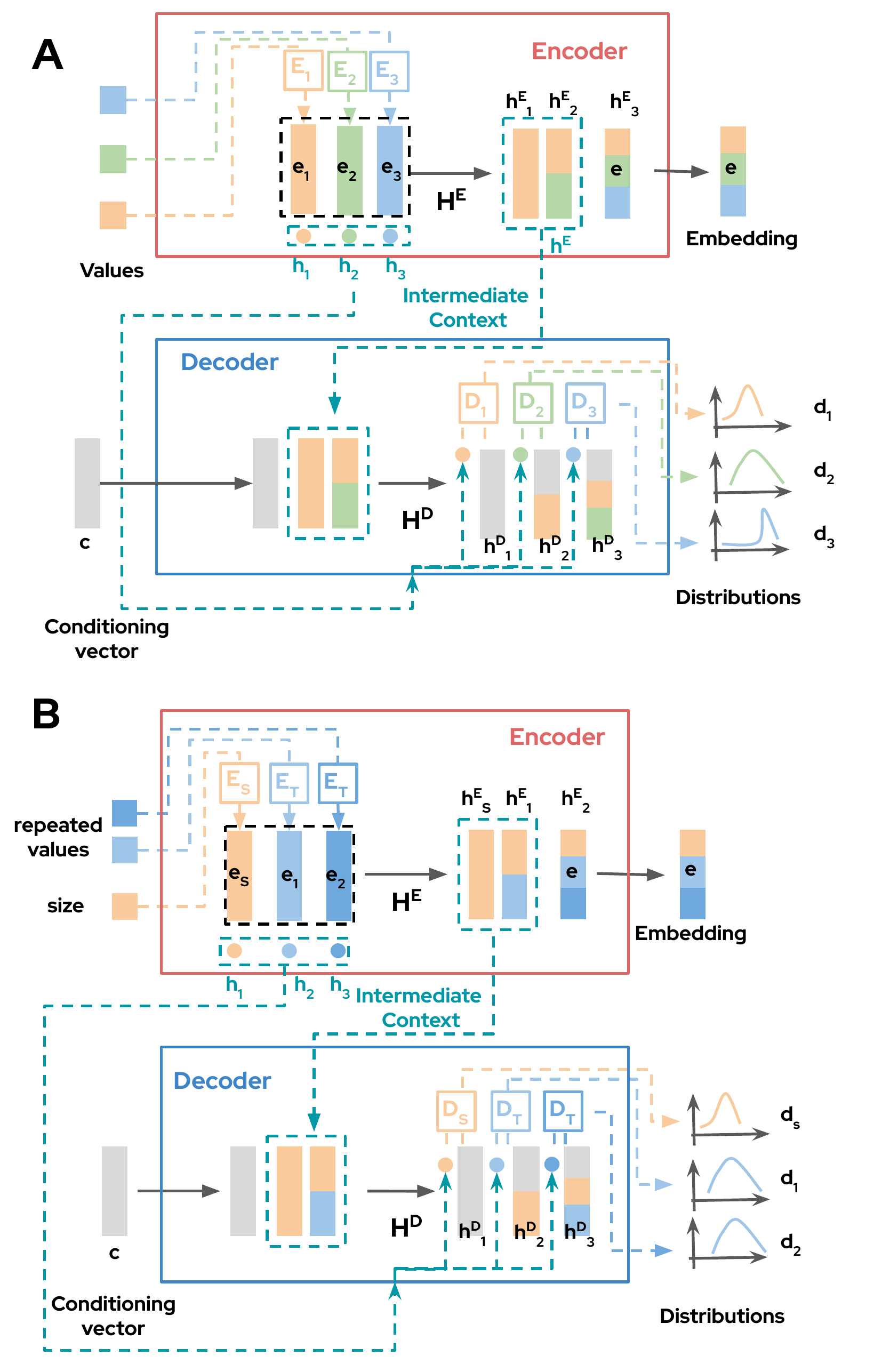}}
\caption{Encoder and decoder for the \textit{struct} (\textbf{A}) and the \textit{list} (\textbf{B}).}
\label{codecs_arch}
\end{center}
\vskip -0.1in
\end{figure}

We can define a \emph{struct codec}
$$C_\text{struct}\left[C_1, C_2,\ldots, C_n\right]$$
by combining $n$ feature codecs $C_k = \left(E_k, D_k, S_k, L_k\right)$, as
illustrated in Fig.~\ref{codecs_arch}-A and Fig.~\ref{struct_sampler} of \ref{annexe_struct}.

The composite encoder $E_\text{struct}$ encodes an observed struct $x = \left(x_1, x_2,\ldots, x_n\right)$ into
an embedding $e \in \mathcal{E} = R^d$ and an intermediate context $h \in \mathcal{H}$
by first encoding each feature $\left(e_k, h_k\right) = E_k\left( x_k \right)$.
A \emph{causal transformer} block $H^E$ (see \ref{transformer}) is then applied to the $e_k$'s,
yielding a list of representations used to build the \emph{intermediate context} and the \emph{struct embedding vector}.
$$\left(h_{1}^{E}, \ldots, h_{n}^{E}\right) = H^E\left(e_1, \ldots, e_n\right)$$
where $h_{k}^{E} \in \mathcal{E} = R^d$.
Since the transformer is causal, we can write:
$$\left(h_{1}^{E}, \ldots, h_{k}^{E}\right) = H^E\left(e_1, \ldots, e_k\right)$$
for all $k$. The vector $h_{k}^{E}$ can be seen as a digest of the embeddings
$e_1, \ldots, e_k$ of the $k$ first features of the struct.
The last digest is used as the \emph{embedding} $e = h_{n}^{E}$ of the struct and
the other digests are packed with the intermediate contexts of the features
to form the \emph{intermediate context} of the struct:
$$h = \left(\left(h_{1}^E,\ldots, h_{n-1}^E\right), \left(h_1, h_2,\ldots, h_n\right)\right)$$
This \emph{intermediate context} is made of two parts: a first part made of the $h_k^E$'s
encoding all the information contained in $x_1,\ldots,x_k$ so that we make sure that
when the distribution of $x_k$ depends on $h_{k-1}^E$ it is conditional on previously considered features in accordance with the chain-rule ;
and a second part with the $h_k$ containing the \emph{intermediate contexts} of the
sub-codecs $C_k$, these will be uninformative for primitive features, but will be useful when the
features are themselves composite types.

The decoder $D_\text{struct}$ takes a conditioning vector $c\in \mathcal{E} = R^d$
and an intermediate context $h \in \mathcal{H}$ and returns a vector of distribution representations:
$$d \in \mathcal{D}\left(\mathfrak{X}\right) = \mathcal{D}\left(\mathfrak{X}_1\right)\times\ldots\times\mathcal{D}\left(\mathfrak{X}_n\right)$$
The conditioning vector is combined with the intermediate context:
$$h = \left(\left(h_{1}^E,\ldots, h_{n-1}^E\right), \left(h_1, h_2,\ldots, h_n\right)\right)$$
using another \emph{causal transformer} $H^D$:
$$\left(h_{1}^D, \ldots, h_{n}^D\right) = H^D\left(c, h_{1}^E,\ldots,h_{n-1}^E\right)$$
A distribution representation is computed for each feature $d_k = D_k\left(h_{k}^D, h_{k} \right)$
and the vector of representations from the struct \emph{distribution representation}:
$$d = \left(d_1,\ldots, d_n\right)$$
The transformer causality implies that the first feature distribution representation only depends on the struct codec's input conditioning vector $c$,
while the second $d_2$ only depends on $c$ and $x_1$, etc.

The sampler $S_\text{struct}$ takes a conditioning context and sequentially samples each feature.
The first feature is sampled as $x_1 \sim S_1\left(H^D\left(c\right)\right)$
and the next features $x_k$ are sampled by encoding each already sampled feature:
$$\left(e_i, h_i\right) = E_i\left( x_i \right)$$
and drawing autoregressively:
$$x_k \sim S_k\left(H^D\left(c, H^E\left(e_1,\ldots,e_{k-1}\right)\right)\right)$$

The loss $L_\text{struct}$ for one observation is computed by applying the loss $L_k$ to
each distribution representation $d_k$ and observation $x_k$.

Note that observations are read feature by feature in a pre-specified order.

\subsubsection{Shuffled Struct Codec}\label{sec:shuffled}

As underlined above, composite codecs rely heavily on the \emph{chain rule} of probability.
This means that, during the training and sampling phases, the order of the features matters
and a model trained with observations in one order will have to be used with that one.

In order to make a codec permutation-invariant,
we randomly \emph{shuffle} the input embeddings $e_1,\dots,e_n$ before they are fed into $H^E$,
and \emph{unshuffle} the outputs of $H^D$ into $h^D_1,\dots h^D_n$ before they are turned into distribution representations
using their respective codecs. More formally, for a given permutation $\sigma$ of $[1,n]$, we use
$$\left(h_1^E,\dots,h_n^E\right) = H^E\left(e_{\sigma(1)},\dots,e_{\sigma(n)}\right)$$
$$\left(h_{\sigma(1)}^D,\dots,h_{\sigma(n)}^D\right) = H^D\left(c, h_1^E,\dots,h_{n-1}^E\right)$$
This technique can be applied several time on each batch without recomputing the input encoding,
yielding a rather efficient data augmentation method.

\subsubsection{List Codec}
\label{sec:list}

An observation $x$ of \emph{list type} is made of a length $m \in \mathbb{N}$
along with a sequence of $m$ values of the same type $\left(m, x_1, x_2,\ldots, x_m\right)$.

We can define a \emph{list codec} by combining a codec for the length and one for the
repeated values:
$$C_\text{list}\left[C_\text{len}, C_\text{val}\right]$$
$C_\text{len}$ can, for instance, be a categorical codec with categories being all possible
lengths. This is illustrated in Fig.~\ref{codecs_arch}-B, and Fig.~\ref{list_sampler} of \ref{annexe_struct}.

The composite encoder $E_\text{list}$ encodes an observed list
$$x = \left(m, x_1, x_2,\ldots, x_m\right)$$
into an embedding $e \in \mathcal{E} = R^d$ and an intermediate context $h \in \mathcal{H}$
by first encoding the list length:
$$(e_\text{len}, h_\text{len}) = E_\text{len}\left( m \right)$$
and encoding the values sequentially:
$$\left(e_k, h_k\right) = E_\text{val}\left( x_k \right)$$
We then apply a \emph{causal transformer} block $H^E$ (see \ref{transformer})
to build the \emph{intermediate context} and the \emph{struct embedding vector}:
$$\left(h_\text{len}^{E}, h_{1}^{E}, \ldots, h_{m}^{E}\right) = H^E\left(e_\text{len}, e_1, \ldots, e_m\right)$$
where $h_\text{len}^{E}$ and $h_k^E \in \mathcal{E} = R^d$.
Similarly to what is done for struct, the last digest is used as the \emph{embedding}:
$$e = h_{m}^{E}$$
of the list and the other digests are packed with the intermediate contexts
to form the \emph{intermediate context} of the list:
$$h = \left(\left(h_\text{len}^E, h_{1}^E,\ldots, h_{m-1}^E\right), \left(h_\text{len}, h_1, \ldots, h_m\right)\right)$$

The decoder $D_\text{list}$ takes a conditioning vector $c\in \mathcal{E} = R^d$
and an intermediate context $h \in \mathcal{H}$ and returns a vector of distribution representations:
$$d \in \mathcal{D}\left(\mathfrak{X}\right) = \mathcal{D}\left(\mathfrak{X}_\text{len}\right)\times \mathcal{D}\left(\mathfrak{X}_\text{val}\right)^m$$
The conditioning vector is combined with part of the intermediate context using another \emph{causal transformer} $H^D$:
$$\left(h_\text{len}^D, h_{1}^D, \ldots, h_{m}^D\right) = H^D\left(c, h_\text{len}^E, h_{1}^E,\ldots, h_{m-1}^E\right)$$
A distribution representation is computed for the length:
$$d_\text{len} = D_\text{len}\left(h_\text{len}^D, h_\text{len}\right)$$
and each value:
$$d_k = D_\text{val}\left(h_{k}^D, h_{k} \right)$$
and the vector of representations from the list \emph{distribution representation}
$$d = \left(d_\text{len}, d_1,\ldots, d_m\right)$$
Note that $d_\text{len}$ only depends on $c$, and $d_k$ depends on $c,m,x_1,\dots,x_{k-1}$.

The sampler $S_\text{list}$ takes a conditioning context, samples the number $m$
of repetitions then sequentially samples $m$ values. The length feature $m$ is sampled as:
$$m \sim S_\text{len}\left(H^D\left(c\right)\right)$$
and the values $x_k$ are sampled by encoding the length:
$$(e_\text{len}, h_\text{len}) = E_\text{len}\left( m \right)$$
and each already sampled feature:
$$\left(e_i, h_i\right) = E_\text{val}\left( x_i \right)$$
and drawing autoregressively:
$$x_i \sim S_\text{val}\left(H^D\left(c, H^E\left(e_\text{len}, e_1,\ldots,e_{i-1}\right)\right)\right)$$

The loss $L_\text{list}$ for one observation is computed by applying the losses $L_\text{len}$ to $d_\text{len}$ and $m$
and $L_\text{val}$ to each distribution representation $d_k$ and observation $x_k$.

\subsubsection{Set Codec}

An observation $x$ of \emph{set type} is a list type (see \ref{sec:list}) whose distribution over its elements is permutation-invariant.
In order to define a \emph{set codec}, we simply use a list codec along with the data augmentation trick of \ref{sec:shuffled}.

\subsection{Mapping data types to codecs}

Since both primitive types and type constructors (Struct, List) can be mapped to codecs
it is straightforward to prove that a dataset whose schema can be described by those
primitive types and constructors can be mapped to a codec too.
For instance, we can define a codec for each column of a table.
Therefore, a single table can be simply seen as a \textit{struct}.
In Tab.~\ref{tab:conversion} of \ref{sc:conv_supp}, we show how we can retrieve the corresponding codec
for a dataset composed of two tables with a one-to-many relationship.
We provide both the \emph{SQL} and the \emph{avro} schema along with the codec.
It illustrates how closely our composite codec system matches the \emph{avro} type system.

\section{Experiments}

In this section, we provide experiments on a number of datasets to show the capacity of our model to generate both simple and hierarchical datasets and assess the quality of the synthetic data.
The model is implemented with \emph{Jax}, see \ref{jax_impl} for more information.
Training and sampling are performed on a single V-100 GPU.

\subsection{Evaluation metrics}
Note that there is not yet a consensus in the machine learning community on a corpus of benchmark datasets
nor on how to evaluate fairly the quality of synthetic data \cite{borisov2021deep}. We focus on three kinds of metrics:

\paragraph{Machine learning utility.} All datasets have a target variable,
for which we use the rest of the variables to perform classification.
For each dataset and model, we generate a synthetic version.
We evaluate the difference in performance on a hold-out set for a classifier trained on the synthetic data and one trained on the real data.
The machine learning performance is measured via the accuracy, F1-score and area under the ROC.
A score closer to zero means that the synthetic data is good enough to be used instead of the real data.
We use 5 different classifiers for each scoring (Logistic Regression, Decision Tree, Mlp classifier,
Random Forest classifier and SVM) and report the average score.

\paragraph{Statistical similarities.} For numerical columns, we compare the Wasserstein distance between the synthetic and real dataset, while we use the Jensen distance for categoricals. To evaluate feature correlations,  we compute the norm of the difference between the pair-wise correlation matrices (real and synthetic).  For a dataset,  the correlation between any two continuous features, corresponds to the Pearson
correlation coefficient while for categorical, we compute the Theil uncertainty coefficient. For a tuple categorical/continuous we compute the correlation ratio. Altogether, this evaluation is the same as in \cite{zhao21} and we borrowed its evaluation code\footnote{\url{https://github.com/Team-TUD/CTAB-GAN/blob/main/model/eval/evaluation.py}}.

\paragraph{Computational efficiency.} We compare the time taken to train each model and to generate data.

\subsection {Standard benchmark datasets}

\paragraph{Datasets.} We test our algorithm on five standard machine
learning datasets with numerical and categorical columns, taken either from the UCI machine learning repository
(Adult, Covertype and Intrusion \cite{dua2017graff} or Kaggle (Credit\footnote{\url{https://www.kaggle.com/mlg-ulb/creditcardfraud}} and Loan\footnote{\url{https://www.kaggle.com/itsmesunil/bank-loan-modelling}}).

\paragraph{Baselines.} We compare our model to 3 state-of-the-art
tabular data generators a gaussian copula model, a variational autoencoder (TVAE) and CTAB-GAN \cite{zhao21},
the last state-of-the-art GAN for tabular data.
For the following results, we retrained the two first models
(the models are obtained from the Synthetic Data Vault\footnote{\url{https://github.com/sdv-dev/SDV/tree/master/sdv/tabular}})
while we directly retrieved the results for the GAN.

\paragraph{Machine Learning Utility.} In Tab.~\ref{benchmark-utility}, we report the averaged metrics over the 5 datasets. Globally, our model outperforms all the other algorithms: the difference in F1 score is 50\% smaller than the other best model while for the AUC it is even 60\%.

\begin{table}[htb!]
\caption{\textbf{Benchmarks utility metrics.} Averaged utility metrics over the 5 datasets. Our model globally outperforms the others for all of them.}
\label{benchmark-utility}
\vskip 0.1in
\begin{center}
\begin{small}
\begin{sc}
\begin{tabular}{lcccr}
\toprule
Model & $\Delta$Accuracy & $\Delta$F1 & $\Delta$AUC \\
\midrule
Copula    & 13.59\%& 0.34& 0.205 \\
CTAB-GAN & 9.83\% & 0.127& 0.117\\
Ours    & \textbf{3.66\%}&  \textbf{0.084}&  \textbf{0.037}\\
TVAE  & 10.11\%& 0.21& 0.099\\
\bottomrule
\end{tabular}
\end{sc}
\end{small}
\end{center}
\vskip -0.1in
\end{table}

\begin{figure}[htb!]
\vskip 0.1in
\begin{center}
\centerline{\includegraphics[width=\columnwidth]{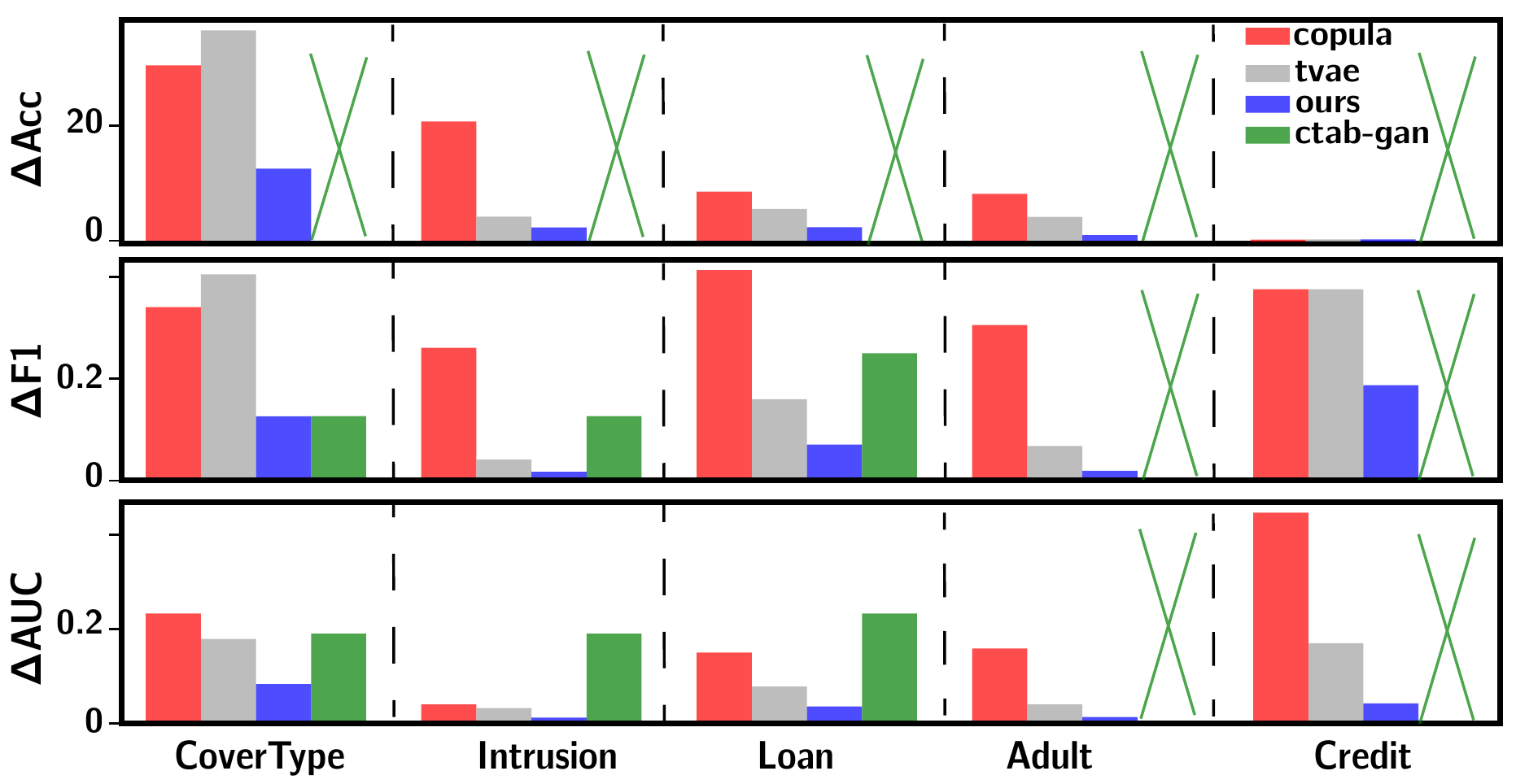}}
\caption{Bar plot of the performances of each model on each dataset.  Each value represents the gap between the score of classifiers trained on the real or synthetic dataset. Hence the smaller the score, the better. Missing values are represented by a cross. Our model (in blue) consistently outperforms all the others for available data.}
\label{auc_benchmark}
\end{center}
\vskip -0.1in
\end{figure}

We further show in Fig.~\ref{auc_benchmark}, the results for each dataset: again, our model consistently achieves the best results.  For datasets with balanced labels such as Adult and Intrusion,  the difference in F1 score and AUC is as small as 4\%. Furthermore, for when the label to be predicted is extremely unbalanced (eg: Credit) TVAE and Copula fail to predict the rare label, which is not the case for our model. We note that two datapoints are missing for CTAB-GAN as they were not explicited in their work (Adult and Credit), hence we cannot fully compare on these datasets. The average score though suggests that our model is at least matching their performance if not beating it.

\paragraph{Statistical similarities}. Results are reported
in Tab.~\ref{benchmark-stats}. For categorical columns, our model strongly outperforms all the others, exhibiting a Jensen distance halved as compared to the second best. For numerical columns,  we report two scores the normalized and un-normalized Wasserstein distance (normalized meaning that all numerical columns are normalized between 0 and 1 before the evaluation).  It is worth noting that for the un-normalized case, the score is strongly influenced by one column of Intrusion(src-bytes) that has a long-tail with few extremely large values (1e8). This completely shifts the distribution if not correctly modeled and masks the performance on all the other datasets (see for example the difference in scores for the copula model). We rather report the un-normalized version because it was the only provided by CTAB-GAN.  In both cases,  though our model largely outperforms all the others.

\begin{table}[!htb]
\caption{\textbf{Benchmarks statistical metrics.}Averaged statistical metrics over the 5 datasets. Wasserstein distance is computed for numerical columns (the un-normalized version is given in brackets) and Jensen distance for categorical. }
\label{benchmark-stats}
\vskip 0.1in
\begin{center}
\begin{small}
\begin{sc}
\begin{tabular}{lcccr}
\toprule
Model & Wasserstein & Jensen & Correlation \\
\midrule
CTAB-GAN & missing (1050)& 0.0697& 2.1\\
TVAE  & 0.01804 (1716)& 0.1459& 2.69 \\
Ours    & \textbf{0.00705 (630)}& \textbf{0.0283}& \textbf{2.03} \\
Copula    & 0.0224(17000)& 0.170& 4.59 \\
\bottomrule
\end{tabular}
\end{sc}
\end{small}
\end{center}
\vskip -0.1in
\end{table}

\paragraph{Computational efficiency.} It takes only a few minutes to train on each dataset and sample with a batch size of 1024. On the other hand, the GAN results are presented for a sub-sample of 50k rows only because computations were too intense, and it took us multiple hours to train and generate the larger datasets with Copulas and TVAE.

\subsection{American Community Survey dataset}

The annual American Community Survey (ACS) is a dataset published by the National Institute of Information and Technology (NIST),  used recently for a differential privacy synthetic generation challenge. It holds one single table with around 1 million rows,  each mapping 35 categorical features of an individual.  Such features contain both physical (sex, age...),  socio-economical (employment status,  income...) and geographical (region, area..) characteristics.  A last column specifies the individual id.
Since the data covers multiple years,  an individual may appear up to 7 times.
\paragraph{Metrics.} The challenge metrics consist in randomly selecting a set of permutations of 4 columns and taking the average total variation norm over the corresponding 4way marginals between the synthetic and real dataset. The score is then remapped from 0 to 1000 (worse to better). Note that such metric in particular does not take into account the consistency across data of the same individual. Therefore,  we also introduce a individual consistency metric,  that accounts for it. In short, we make checks on the sex, age,  race and education across records of each individual and report the percentage of inconsistencies. A more detailed explanation is given in \ref{sec:acs_supp}.

\paragraph{Record-level dataset.}If we do not consider individual consistency, the dataset can be modeled by a \emph{struct of categorical}.  In this case,  our model outperforms both standard single-table models (CTGAN, CopulaGAN) and specially designed models crafted by the challenge participants (SDNIST avg and top) (see Tab. \ref{tab:acs-record}).\footnote{On the one hand, this comparison should be taken with a grain of salt as the SDNIST rows correspond to a privacy setting of ($\epsilon=10$). On the other, $\epsilon=10$ formally provides a weak protection against membership inference attack, and participants had access to a public dataset with very similar marginals that allowed them to make correct assumptions on the private one.}

\paragraph{User-level dataset.}However, we can take into account individual consistency by considering a user-level model as shown in Fig.\ref{fig:acs_schema}. In this case, our synthetic data has 88\% of the users with all consistent records for both age, sex, race and education. Furthermore, the score on the marginals is comparable with the previous one. We also train an instance of HMA1, though only on a sample of 100k rows for computational reasons.  For a fair comparison, we train our model on the same sub-sample.  While our synthetic dataset is still robust (76\% of consistency), the HMA1 sample is quite weak (only 28\%). Furthermore, the marginals score is also 15\% lower.
\begin{table}[htb!]
\caption{\textbf{ACS results.} The marginals column refers to the $4-$way marginal score while consistency to the individual consistency metric.  The top-half of the table reports results for algorithms that do not consider individuals and so this metric is not calculated.}
\label{tab:acs-record}
\vskip 0.1in
\begin{center}
\begin{small}
\begin{sc}
\begin{tabular}{lccr}
\toprule
Model  & marginals & Consistency \\
\midrule
SDNIST (avg) &893 &  \\
SDNIST (top) & 901 &\\
CTGAN & 738 \\
CopulaGAN & 773 \\
Ours (Record) &  \textbf{903} & \\
\midrule
Ours (User) &  \textbf{898} & 88\% \\
HMA1 (100k) & 659 & 28\% \\
Ours (100k) & \textbf{769} & \textbf{76\%} \\
\bottomrule
\end{tabular}
\end{sc}
\end{small}
\end{center}
\vskip -0.1in
\end{table}

\paragraph{Computational efficiency.} As stated above, we were unable to generate the full dataset with HMA1 (generating 100k rows took a couple hours), while our model has been trained and generated the data a few minutes (less than 8 minutes for the whole dataset).

\subsection{Netflix Prize\label{sec:netflix}}

From 2006 to 2009, Netflix held a competition where the goal was to improve its movie rating predictions \cite{Bennett2007TheNP}.  Even though the dataset was anonymized,  researchers \cite{narayanan2007break} were able to re-identify users by matching with IMDb ratings.  Following privacy and legal concerns, the competition was discontinued.  The dataset contains around 100M reviews from 500,000 users about 17,700 movies.  It is organized in two tables (\emph{movies} and \emph{ratings}) linked by unique movie identifiers.

\begin{figure}[htb]
    \label{fig:netflix_schema}
    \centering
    \includegraphics[width=0.7\columnwidth]{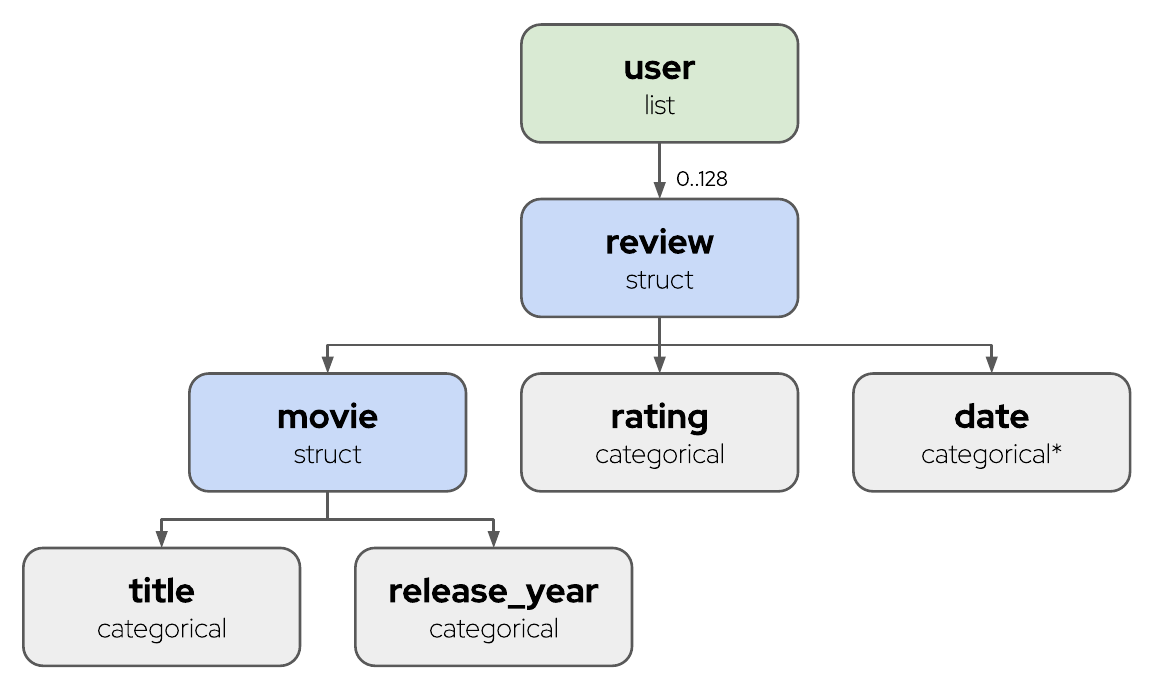}
        \caption{\textbf{Netflix dataset codec.}  A user is associated to a list codec containing repeated values of a struct codec (the review). Each review has a categorical codec for rating and date and a struct codec (movie) containing a title and a release year (both categoricals).}
\end{figure}

\paragraph{Experimental setup.} The whole dataset can be interpreted as a hierarchical dataset as shown in Fig.~\ref{fig:netflix_schema}.  Such architecture in particular allows the model to learn correlations between any of the movie features (\emph{title}, \emph{release\_year}) and the user preferences. We chose to limit ourselves to 4500 movies and 128 reviews per user. This reduces the number of reviews to $\approx 25M$.  Note that it could not be possible to generate the same reduced dataset with HMA1 again for computational reasons.

\paragraph{Machine Learning Utility.}We use the Surprise library \footnote{\url{http://surpriselib.com}},  a recommender system library that provides built-in algorithms to predict the user movie rating based on the other attributes.  In particular, it provides $SVD$, the algorithm that won the competition. The results, given in Tab. \ref{tab:netflix}, show the RMSE on an hold out portion of the real dataset when the algorithm is trained either on the rest of the real data or on the synthetic ones.  We only have $\approx 10$\% of difference,  which shows the robustness of our synthetic sample.
\begin{table}[!htb]
\caption{\textbf{Netflix results.} Different algorithms are trained to predict the ratings either on the real dataset or the synthetic one.  The RMSE of the prediction on a hold-out part of the real dataset is reported.}
\label{tab:netflix}
\vskip 0.1in
\begin{center}
\begin{small}
\begin{sc}
\begin{tabular}{lccr}
\toprule
Algorithm & Train Real & Train Synthetic\\
\midrule
  NormalPredictor & 1.48 & 1.47\\
  BaselineOnly & 0.94 & 1.02\\
  SVD & 0.92 & 1.02 \\
  SlopeOne & 0.94 & 1.08\\
  CoClustering & 0.96 & 1.08\\
\bottomrule
\end{tabular}
\end{sc}
\end{small}
\end{center}
\vskip -0.1in
\end{table}

\begin{figure}[!htb]
\vskip 0.1in
\begin{center}
\centerline{\includegraphics[width=\columnwidth]{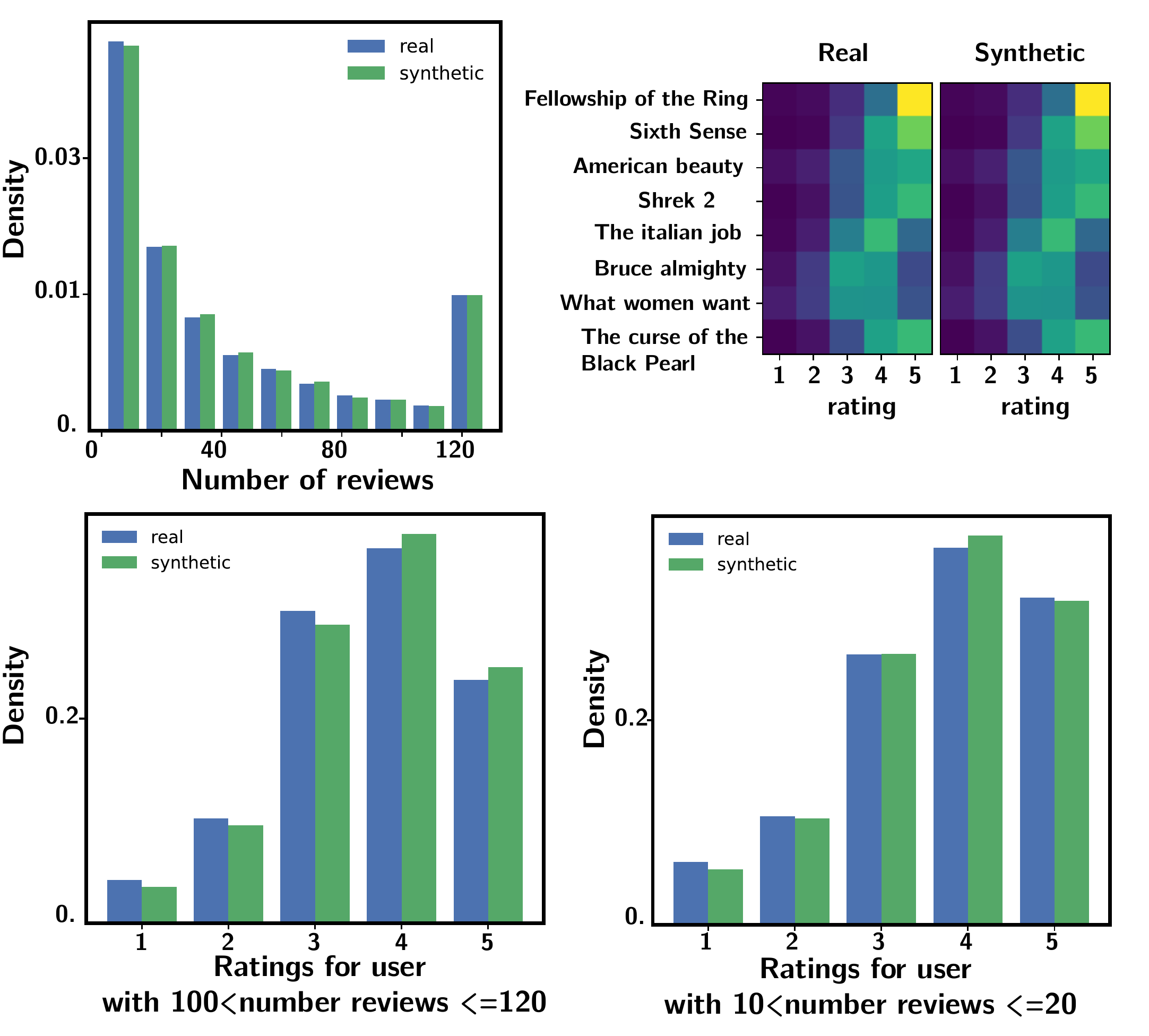}}
\caption{Plots of different conditional distributions for the real and synthetic dataset:
the number of reviews per user, the rating for a number of films and the distribution of ratings filtered for users with few reviews and many reviews
}
\label{netflix_marginal}
\end{center}
\vskip -0.1in
\end{figure}

\paragraph{Statistical similarities.} We handpicked a set of non trivial marginals and conditional distributions.
These distributions are complex either because the conditioning is not explicitly modeled by our hierarchical model (for instance, the number of reviews) or because they are very sparse.  As shown in Fig.\ref{netflix_marginal} such distributions are still well captured by the synthetic data.

\paragraph{Computational efficiency.} We could not generate the sampled dataset with HMA1, while our model trains on 24M ratings in less than 5 minutes (1 epoch with a batch size of 512).

\section{Conclusion and future works}
In this paper, we presented a new framework for synthesizing complex data types.
Our approach allows us to tackle complex data in a systematic way,
mapping in particular \emph{avro} data schema -- a widespread format in the industry.
Moreover, in all our benchmarks, we consistently top the state-of-the art.
Besides, our algorithm is fast and scales well to large datasets (see Netflix prize).
Finally, our abstract definition of a codec is general enough that many other generative models can be used as codecs.

A natural development of this work would be to add more codecs, using pre-trained models:
for instance an \emph{image codec} could be built out of a conditional GAN such as \emph{BigGAN} and an image encoder such as \emph{ResNet}.
The GAN generator would be used for decoding and sampling and the discriminator for the loss.
Something analogous could be done for text, using a standard text transformer such as GPT2.

Finally, although privacy protection is a motivation of this work, it is worth noting that synthetic data generation alone does not guarantee such protection.
A standard way of making sure synthetic data preserve privacy is to train the generative model with \emph{differential privacy} guarantees \cite{dwork2014algorithmic,abadi2016deep}.
We show some preliminary encouraging results in \ref{annexe_dp}.
In addition, our model's ability to  leverage existing pre-trained models could be a real asset in the area of privacy preserving data generation and analysis, opening many new possibilities.

\bibliography{metalearner}
\bibliographystyle{alpha}

%%%%%%%%%%%%%%%%%%%%%%%%%%%%%%%%%%%%%%%%%%%%%%%%%%%%%%%%%%%%%%%%%%%%%%%%%%%%%%%
%%%%%%%%%%%%%%%%%%%%%%%%%%%%%%%%%%%%%%%%%%%%%%%%%%%%%%%%%%%%%%%%%%%%%%%%%%%%%%%
% APPENDIX
%%%%%%%%%%%%%%%%%%%%%%%%%%%%%%%%%%%%%%%%%%%%%%%%%%%%%%%%%%%%%%%%%%%%%%%%%%%%%%%
%%%%%%%%%%%%%%%%%%%%%%%%%%%%%%%%%%%%%%%%%%%%%%%%%%%%%%%%%%%%%%%%%%%%%%%%%%%%%%%
\newpage
\appendix
\onecolumn

\section{Additional Technical information}

\subsection{Struct and List samplers} \label{annexe_struct}

\begin{figure}[!htb]
\vskip 0.1in
\begin{center}
\centerline{\includegraphics[width=0.8\columnwidth]{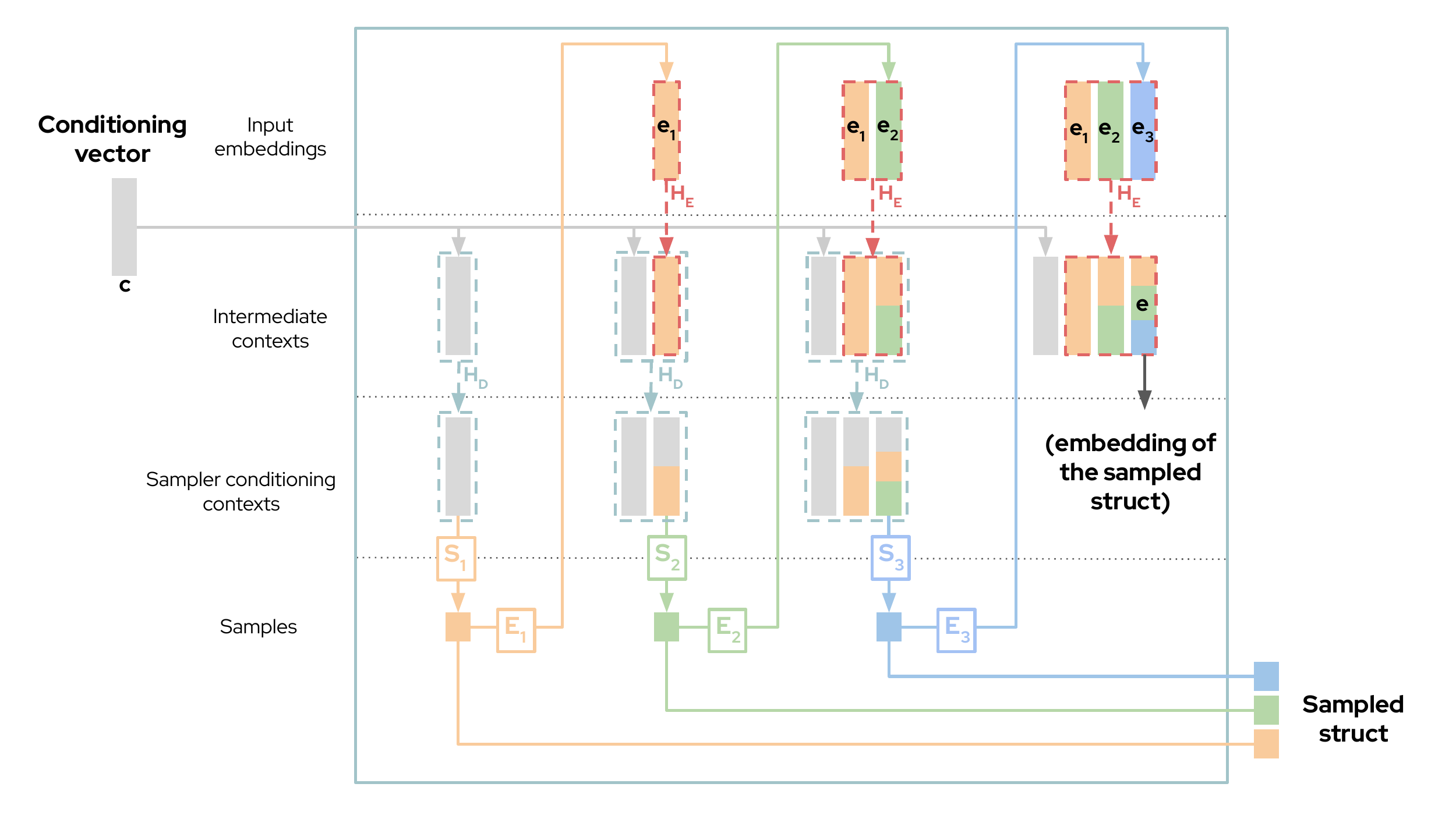}}
\caption{Sketch of the struct sampler}
\label{struct_sampler}
\end{center}
\vskip -0.1in
\end{figure}

\begin{figure}[!htb]
\vskip 0.1in
\begin{center}
\centerline{\includegraphics[width=0.8\columnwidth]{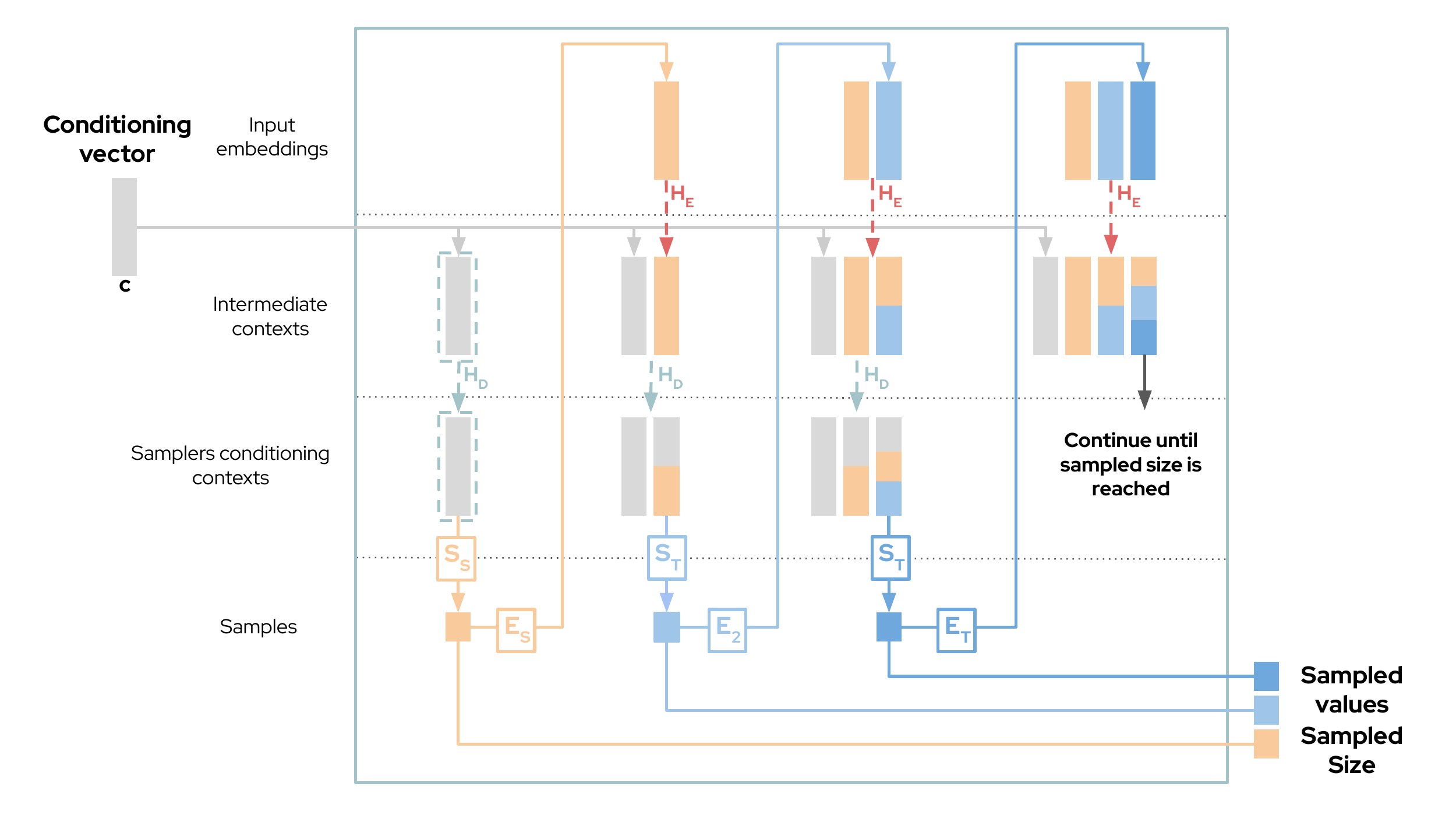}}
\caption{Sketch of the list sampler}
\label{list_sampler}
\end{center}
\vskip -0.1in
\end{figure}

\subsection{Data structure-codec conversion}\label{sc:conv_supp}

\newpage
\begin{landscape}

\begin{table*}[htb!]
\caption{\textbf{Data-structure to codec conversion.} We provide an example of how two tables with a relational link would be written either in SQL or Avro and then what would be our associated codec.}
\label{tab:conversion}
\vskip 0.1in
\begin{center}
\begin{tabular}{lcccr}
\toprule
SQL Format & Avro schema & Codec \\
\midrule
\begin{lstlisting}[
           language=SQL,
           showspaces=false,
           basicstyle=\ttfamily,
           numbers=none,
        ]
CREATE TABLE Users(
Age Float
Sex Enum
Id Enum);

CREATE TABLE Transaction(
Place Enum
Price Float
UserId Enum
FOREIGNKEY (UserId)
REFERENCES Users(Id));
\end{lstlisting} & \begin{lstlisting}[
           language=python,
           showspaces=false,
           basicstyle=\ttfamily,
           numbers=none,
           tabsize=1
        ]
{
  "type": "record",
  "name": "User",
  "fields": [
    {
      "name": "Age",
      "type": "float"
    },
    {
      "name": "Sex",
      "type": "enum"
    },
    {
      "name": "transactions",
      "type": "array",
      "items": {
        "type": "record",
        "name": "transaction",
        "fields": [
          {
            "name": "Place",
            "type": "enum"
          },
          {
            "name": "Price",
            "type": "float"
          }
        ]
      }
    }
  ]
}
\end{lstlisting} & \shortstack[l]{User: $C_{\text{struct}}$[\\ \;Age: $C_{\text{num}}$, \\ \;Sex: $C_{\text{cat}}$, \\ \;Transactions: $C_\text{list}$[\\ \; \; \;Transaction: $C_{\text{struct}}$[\\ \qquad Place: $C_{\text{cat}}, $\\ \qquad Price: $C_{\text{num}}$]\\ \qquad \qquad ] \\ \qquad]} \\\\

\bottomrule
\end{tabular}
\end{center}
\vskip -0.1in
\end{table*}
\end{landscape}

\subsection{Transformer implementation}\label{transformer}

The transformer was initially made of multiple blocks based on the GPT2 architecture: each block consisted namely in one layer of layer-normalization, one of multi-head self-attention with a causal mask, another layer normalization and a dense layer. So that the output $y$ can be written as a function of the input $x$ as:
\begin{align}
\tilde{x}=\text{LayerNorm}_1(x) \\
\tilde{y}=\text{CausalSelfAttention}(\tilde{x})+x \\
y=\text{Dense}(\text{LayerNorm}_2(\tilde{y}))+\text{LayerNorm}_2(\tilde{y}) \\
\end{align}
In our case, since the transformer is not very deep (typically 2 blocks are enough), we found experimentally that removing the layer normalization and the dense did not alter the results. This is consistent with other works \cite{geva2021transformer}. Everything presented in this paper is done with a reduced transformer that computes:
\begin{align}
y=\text{CausalSelfAttention}(x)+x
\end{align}
This has the advantage of strongly reducing the number of parameters.

\subsection{\texttt{Jax} implementation}\label{jax_impl}

\par\textbf{Codec instantiation.} Our codecs are implemented in \texttt{Jax} and \texttt{Flax} and heavily depend on its ability to transform pure functions. In particular, automatic vectorization using \texttt{vmap} allows to write codecs as independent blocks, without worrying about batch and feature dimensions, which could become quite cumbersome in the hierarchical cases.  All the codecs of a composite codec can directly be instantiated from dataclasses describing their type,  in a declarative fashion. The end-user does not have to worry about pipework.  For instance, the hierarchical Netflix model of section \ref{sec:netflix} (described in Fig. \ref{fig:netflix_schema}) can be instantiated using type definition of listing \ref{lst:netflix}. In particular, the parameters of the \texttt{movie} codec instantiated from the corresponding type are shared, allowing it to effectively learn the \texttt{release\_year} and \texttt{title} associations of all movies in a single sub-codec\footnote{Learning the relationship between titles and release years may seem unnecessary, however this allows us to include both information in the embedding of a movie or that of a user}.

\lstset{language=Python,
        basicstyle=\ttfamily\small,
        stringstyle=\color{gray},
        showstringspaces=false,
        identifierstyle=\color{darkgray},
        keywords=[2]{pow},
        keywordstyle=[2]{\color{orange}},
        numbers=left
}
\begin{figure}[!htb]
    \begin{lstlisting}[caption=Netflix type definition,label=lst:netflix]
BatchType("dataset",
  ShuffledListType("user",
    StructType("review",
      {
        "movie": StructType("movie",
          {
            "release_year": CategoricalType("release_date",94),
            "title": CategoricalType("title", 4500),
          }
        ),
        "year": CategoricalType("year", 7),
        "rating": CategoricalType("rating", 5),
      }
    ),
  ),
)
    \end{lstlisting}
\end{figure}

\paragraph{Training} During training, the model is directly fed a single Jax's \texttt{pytree} representing the element that is learnt from. A \texttt{pytree} is a tree-like structure of container-like Python objects (\texttt{dict}, \texttt{tuple}, \texttt{list},...) where the leaves are tensors. Vectorizing (\texttt{vmap}) directly operates on leaves (tensors), traversing the nodes (python containers).

In order to allow our model to process batches, we simply define a \texttt{BatchType}, representing an array of independent elements with a fixed size (corresponding to the batch size). Overall, codecs can internally expect to receive a single element of the type they are encoding ; the rest is handled using vectorization (\texttt{vmap}). Our codecs thus internally expect the following inputs:

\begin{itemize}
    \item \texttt{CategoricalCodec} expects a single integer value (an integer tensor of dimension 0), representing the index of the category of the input.
    \item \texttt{StructCodec} and its shuffled variant expect to receive a single Python \texttt{dict} where the keys are the sub-type names and the values are the values (as \texttt{pytrees}) of the corresponding sub-types.
    \item \texttt{ListCodec} and its shuffled variant expect a Python \texttt{tuple} \texttt{(size, values)}, where \texttt{size} is the length of the input sequence (an integer tensor of dimension 0), and \texttt{values} is another \texttt{pytree} corresponding to its repeated sub-type, where the first dimension of the leaves correspond to elements of the list. \texttt{size} can be lower than the number of elements that are actually received (which is itself fixed) by masking the final elements. This allows to efficiently vectorize and compile the model.
    \item \texttt{BatchCodec} expects a \texttt{pytree} corresponding to the repeated sub-type (usually the type of a user) where the first dimension of the leaves correspond to the batch dimension.
\end{itemize}

For instance, the type of input of the Netflix model is given by listing \ref{lst:input}. The input type is in fact a transposition of the type hierarchy, as "tensorizable" (arrays) dimensions are put after the "non-tensorizable" ones (dict, tuples).

\begin{figure}[!htb]
    \begin{lstlisting}[caption=Netflix input type, label=lst:input]
(
  np.ndarray[BATCH_SIZE],
  {
   "movie": {
      "release_year": np.ndarray[BATCH_SIZE, REVIEW_SIZE],
      "title": np.ndarray[BATCH_SIZE, REVIEW_SIZE]
   },
   "year": np.ndarray[BATCH_SIZE, REVIEW_SIZE],
   "rating": np.ndarray[BATCH_SIZE, REVIEW_SIZE],
  }
)
    \end{lstlisting}
\end{figure}

\section{Additional Experiment details}

For all the experiments, we use a transformer with 2 blocks and 8 heads per block. The embedding dimension is set to 64.

\subsection{American Community Survey (ACS) Dataset \label{sec:acs_supp}}

\begin{figure}[!htb]
    \caption{Record-level (left) and user-level (right) ACS dataset schema}
    \label{fig:acs_schema}
    \centering
    \includegraphics[width=0.4\columnwidth]{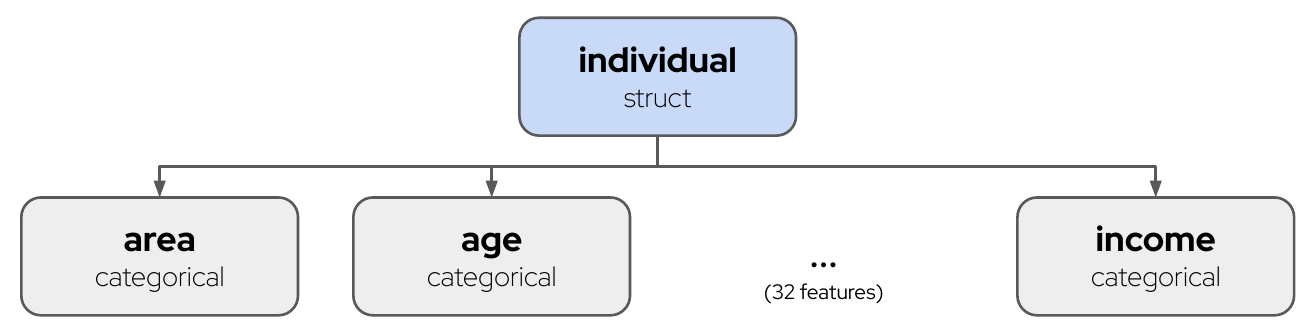}\hspace{1cm}\includegraphics[width=0.4\columnwidth]{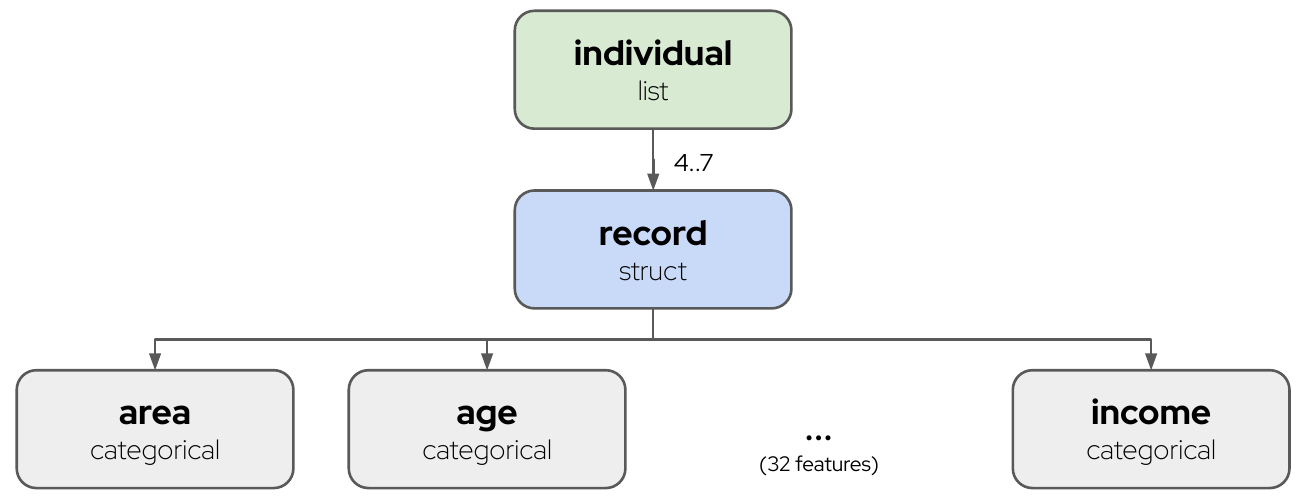}
\end{figure}

\paragraph{Training.} The codec corresponding to each model (record-level and user-level) is represented in Fig. \ref{fig:acs_schema}. Our models only have 31k to 45k parameters, which is considerably below the number of records or individuals. Training is done on a single V100 and takes from 4 to 20 seconds per epoch using a batch size of 1024 to 4096. Sampling a dataset as large as the original one takes around 20 seconds\footnote{Those times are reported without Jax's jit compilation time}.

\par\textbf{Individual consistency metric.} The original ACS dataset has been built with some explicit constraints between records of an individual.  Specifically:
\begin{itemize}
\item the \emph{sex}, \emph{race} and \emph{hispan} attributes are fixed traits in the original dataset. This constrained is referred as \emph{Sex/Race/Hispan};
\item each individual is supposed to have at most one record per year in the original dataset.  This constrained is referred as  \emph{Year};
\item the level of education should increase monotonically. This constraint is referred as \emph{Education};
\item for each record, the birthday of the individual can be inferred and should be constant across the records. This constraint is referred as \emph{Birthday};
\end{itemize}
For each synthetic dataset, we report the fraction of individuals that have a constrained violated.
Note that those constraints are enforced in the original dataset but are only inferred from observations by HMA1 and our models. The results indicate that our model is able to learn and synthesize a variety of constraints in the dataset, with almost perfect accuracy on the full dataset for the simpler ones.

\begin{table*}[!htb]
    \caption{Individual Consistency metrics}
    \label{tab:acs2}

    \centering{}%
    % \begin{sc}
    \begin{tabular}{lccccr}
    \toprule
    Model &  Sex/Race/Hispan & Year & Education & Birthday \\
    \midrule
    HMA1 (100k) & 65\% & 7\% & 32\% & 6\% \\
    Ours (100k)  & \textbf{97\%} & \textbf{60\%} & \textbf{84\%} & \textbf{61\%} \\
    \midrule
    Ours (full) & 99\% & 71\% & 92\% & 89\% \\
    \bottomrule
    \end{tabular}
    % \end{sc}
\end{table*}

\subsection{Netflix Prize \label{sec:netflix_supp}}

The code corresponding to the Netflix dataset has 177k parameters, much lower than the number of reviews (25M) or even the number of users (500k).  It is trained on a single V100 with batch sizes of 512. Training a single batch takes around 4 minutes and the model quickly converges after 3 or 4 epochs, suggesting that it could be suited to handle a lot more users.

%%%%%%%%%%%%%%%%%%%%%%%%%%%%%%%%%%%%%%%%%%%%%%%%%%%%%%%%%%%%%%%%%%%%%%%%%%%%%%%
%%%%%%%%%%%%%%%%%%%%%%%%%%%%%%%%%%%%%%%%%%%%%%%%%%%%%%%%%%%%%%%%%%%%%%%%%%%%%%%

\section{Differential Privacy}\label{annexe_dp}

In this section, we provide some preliminary work to show the promises of our model trained with differential privacy.  In fact,  the model can be easily trained using the standard algorithm of DP-SGD \cite{dpsgd}. In particular,  it is very convenient to work with DP, since the target entity to protect directly corresponds to a top-level compound type (usually a user or an individual) and so it always appears only once in a batch.  Differential private training is performed with the \emph{Optax}\footnote{\url{https://github.com/deepmind/optax}} library.

\paragraph{ACS Dataset.} We train our model with differential privacy with a budget of $\epsilon=1$ and $\delta=2.5\times 10^{-5}$. This corresponds to 5 epochs of DP-SGD with a noise multiplier of 1.08 and batch size of 1024 users. We take a small clipping threshold of $C=10^{-3}$ (see \ref{tab:acsdp}).  We note that our model without pre-training and strong optimizations, is competitive with the average participant, while also modeling user-level correlations and without any assumption on the dataset other than its schema, i.e our model was not pre-trained on a similar dataset with public data.

\begin{table}[htb!]
    \caption{DP results for ACS dataset ($\epsilon=1,\delta=2.5\times10^{-5}$)}
    \label{tab:acsdp}
    \vskip 0.1in
    \begin{centering}
    \begin{tabular}{lcr}
        Model & 4 way marginal \\
    \midrule
    SDNIST (avg) & 842 \\
    SDNIST (top) &\textbf{865} \\
    \midrule
    Our (user-level) & 844 \\
    \bottomrule
    \end{tabular}
    \par\end{centering}
    \end{table}

\paragraph{Netflix Prize}
We also trained our model for a single epoch with DP-SGD, using the same batch size (512). We picked a clipping threshold of $C=10^{-3}$ and set the noise multiplier to 0.92 so to spend a budget of $(\epsilon=1,\delta=1/n)$.  Without any fine-tuning, we obtain a loss in the best score of only 15\% (comparing to SVD on the real dataset) which looks rather encouraging.
\begin{table}[htb!]
\caption{ We add a column with the DP results to the Tab.}
\label{tab:netflixdp}
\vskip 0.1in
\begin{center}
\begin{small}
\begin{sc}
\begin{tabular}{lccr}
\toprule
Algorithm & Train Real & Train Synthetic & Train DP Synthetic\\
\midrule
  NormalPredictor & 1.48 & 1.47 & 1.50\\
  BaselineOnly & 0.94 & 1.02 & 1.09 \\
  SVD & 0.92 & 1.02 &1.10 \\
  SlopeOne & 0.94 & 1.08 &1.10\\
  CoClustering & 0.96 & 1.08 &1.10\\
\bottomrule
\end{tabular}
\end{sc}
\end{small}
\end{center}
\vskip -0.1in
\end{table}
\end{document}